\documentclass[conference]{IEEEtran}
\usepackage[top=54pt, left=54pt, right=54pt, bottom=54pt]{geometry}

\IEEEoverridecommandlockouts
\usepackage{cite}
\usepackage{amsmath,amssymb,amsfonts}
\usepackage{algorithmic}
\usepackage{graphicx}
\usepackage{textcomp}
\usepackage{xcolor}
\usepackage{yhmath}
\def\BibTeX{{\rm B\kern-.05em{\sc i\kern-.025em b}\kern-.08em
    T\kern-.1667em\lower.7ex\hbox{E}\kern-.125emX}}

\usepackage{array}
\usepackage[ruled,linesnumbered]{algorithm2e}

\usepackage{textcomp}
\usepackage{stfloats}
\usepackage{url}
\usepackage{verbatim}
\usepackage{graphicx}
\usepackage{cite}
\usepackage{subfigure}
\usepackage{caption}
\usepackage{multirow}
\usepackage[colorlinks,
            linkcolor=blue,
            anchorcolor=blue,
            citecolor=blue,
            urlcolor=blue]{hyperref}

\begin{document}


\title{GA-TEB: Goal-Adaptive Framework for Efficient Navigation Based on Goal Lines}

\author{Qianyi Zhang$^1$, Wentao Luo$^2$, Ziyang Zhang$^2$, Yaoyuan Wang$^2$, and Jingtai Liu$^{1,\dagger}$,~\IEEEmembership{Senior Member, IEEE}
\thanks{${^1}$ Institute of Robotics and Automatic Information System, Nankai University, Tianjin, China. 
${^2}$ Huawei 2012lab, China. 
$^{\dagger}$ Corresponding author: Jingtai Liu, liujt@nankai.edu.cn. This work is supported by the National Natural Science Foundation of China under Grant 62173189. 
}
}





\maketitle

\begin{abstract}
In crowd navigation, the local goal plays a crucial role in trajectory initialization, optimization, and evaluation. Recognizing that when the global goal is distant, the robot's primary objective is avoiding collisions, making it less critical to pass through the exact local goal point, this work introduces the concept of goal lines, which extend the traditional local goal from a single point to multiple candidate lines. Coupled with a topological map construction strategy that groups obstacles to be as convex as possible, a goal-adaptive navigation framework is proposed to efficiently plan multiple candidate trajectories.
Simulations and experiments demonstrate that the proposed GA-TEB framework effectively prevents deadlock situations, where the robot becomes frozen due to a lack of feasible trajectories in crowded environments. Additionally, the framework greatly increases planning frequency in scenarios with numerous non-convex obstacles, enhancing both robustness and safety.
\end{abstract}

\section{Introduction}
With the consistent advancement of robot technology, the mobile robot navigation framework has gradually matured~\cite{s0,s1}. After a global goal point is given, the global planner searches for a coarse path within the global map and provides a local goal point, while the local planner searches for a fine trajectory within the real-time updated local map, ending at the local goal point~\cite{s2}. 

\begin{figure}[thb]
    \centering
    \includegraphics[width=3.4in]{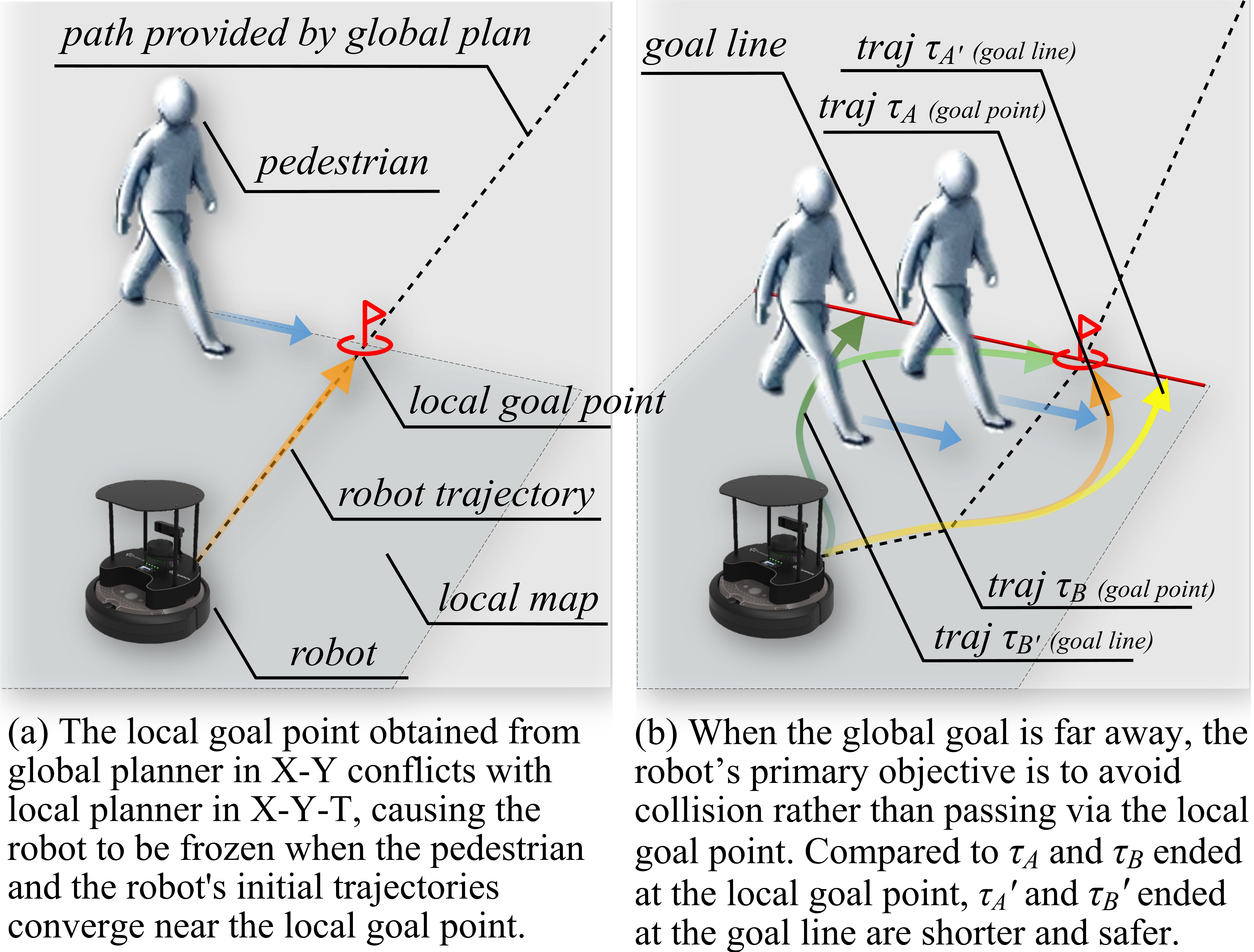}
    \caption{Issues of local goal in classic navigation framework.}
    \label{problem}
\end{figure}

However, in complex dynamic environments, the local goal point may lead to erroneous guidance~\cite{e5}. 
Fig.\ref{problem}(a) illustrates a freezing problem~\cite{freezing} caused by the local goal point. To save time when searching for a coarse path over a large global map, the global planner (such as the widely-used A$^*$ algorithm~\cite{Astar}) typically searches only in the spatial dimension, thus providing a local goal point reflecting only the current positions of obstacles. However, since the subsequent local planner operates in the spatiotemporal dimension, it considers the future states of dynamic obstacles or moving pedestrians. This mismatch can lead to situations where the endpoints of all trajectories overlap with the future positions of obstacles, causing an unavoidable collision regardless of how the trajectory is optimized. Ultimately, a local deadlock occurs, where the robot becomes frozen, unable to find a viable trajectory forward. This issue is particularly common in crowded environments, reducing both safety and efficiency.
Moreover, since all trajectories terminate at the same local goal point, many of them may involve unnecessary detours, which prolongs the trajectory optimization process and leads to incorrect trajectory evaluations. Fig.\ref{problem}(b) illustrates this situation with two trajectories, $\tau_A$ and $\tau_B$, both ending at the local goal point. Trajectory $\tau_A$ detours in front of the pedestrians, while $\tau_B$ detours behind them. In time-optimal navigation, the shorter trajectory $\tau_A$ would typically be preferred. However, an intuitive understanding suggests that when the global goal is far away, the robot’s primary objective is to avoid pedestrians and obstacles, making it less critical to pass through the exact local goal point. Therefore, compared to trajectories $\tau_A$ and $\tau_B$, $\tau_A'$ and $\tau_B'$ represent more reasonable optimization results, as they avoid unnecessary detours, resulting in shorter lengths and quicker optimizations. More importantly, while $\tau_A$ is preferred over $\tau_B$ due to the latter's unnecessary detour, among all the four trajectories, $\tau_B'$ is the best choice due to its shorter length. 

Based on the above observations, this work proposes the concept of a \textit{goal line} (shown in red in Fig.\ref{problem}b), which extends the traditional local goal from a single point to multiple lines. This allows the endpoint of the trajectory to slide along its line during trajectory initialization and optimization, enabling the robot to reach the global goal more efficiently and safely.

Goal lines are particularly effective in crowded scenarios, in alignment with another contribution of this work: a max-convex topological map construction strategy. Crowd navigation has long been a key challenge for mobile robots, leading to various proposed approaches. Social force assumes that obstacles exert repulsive forces on the robot, influencing its motion~\cite{a0}, with variants incorporating group dynamics~\cite{a1}, human intentions~\cite{a2}, and comfort models~\cite{a3}. Velocity obstacle analyzes pedestrian motion in velocity space to find feasible velocities~\cite{b0,b1}, with variants incorporating game theory~\cite{b2}, optimization~\cite{b3}, and protected zones~\cite{b4}. Limit cycle focuses on the nearest obstacle, triggering avoidance maneuvers~\cite{c0}, with variations considering human gaze~\cite{c1} and potential fields~\cite{c2}. Reinforcement learning and deep learning methods use raw perception data~\cite{d0,d1} or estimated object states~\cite{d2,d3} to generate actions. However, as these methods typically do not offer alternative strategies~\cite{e1}, they often lack robustness.

For methods offering alternative strategies, Time-Elastic Band (TEB)~\cite{e2,e3} is a representative work, using Voronoi graphs or probabilistic roadmaps for trajectory initialization, followed by optimization under G2O~\cite{g2o} framework. Variants incorporate topological structures~\cite{e4,e5} and heuristic search strategies~\cite{e6,e7} to accelerate the process. However, an existing issue is that low-quality initialized trajectories can lead to prolonged optimization or convergence to local optima, especially in scenes with non-convex obstacles~\cite{e8}. Building upon TEB, a max-convex topological map construction strategy is proposed in this work, grouping adjacent obstacles and introducing a group-level Voronoi graph to enhance trajectory initialization and optimization, thereby improving navigation robustness and safety.

\begin{figure*}
    \centering
    \includegraphics[width=\linewidth]{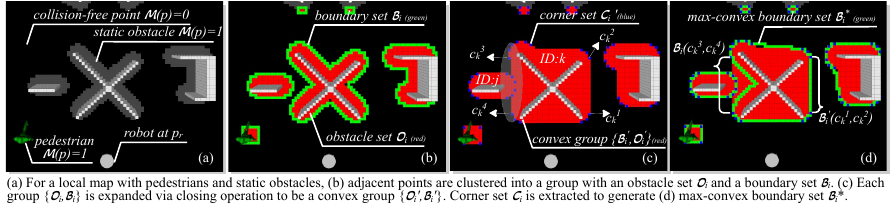}
    \caption{Illustration of obstacle group cluster with max-convex boundary. }
    \label{max_boundary}
\end{figure*}
In summary, a framework, GA-TEB, for efficient crowd navigation is proposed with the following contributions:
\begin{itemize}
    \item A max-convex topological map construction strategy is designed to analyze scenes and initialize trajectories. 
    \item A concept of goal lines is proposed to enhance trajectory quality during initialization and optimization. 
    

\end{itemize}

\section{Topological Map Construction}

\subsection{Obstacle Group Cluster with Max-convex Boundary}

The robot observes its surrounding environment and maintains a local map $\mathcal{M}$ where $\mathcal{M}(p)=1$ indicates that the point $p$ is occupied by pedestrians or static obstacles, and $\mathcal{M}(p)=0$ indicates that $p$ to be collision-free (see Fig.\ref{max_boundary}a). Building on our previous work~\cite{e5}, adjacent obstacle points can be clustered into a group consisting of an obstacle set $\mathcal{O}_i$ and an ordered boundary set $\mathcal{B}_i$ (see Fig.\ref{max_boundary}b):
\begin{align}
    \mathcal{O}_i = \{p  |_{ \exists p': \; p' \in \mathcal{O}_i \; \& \; ||p'-p||_{\infty}=1 \; \& \; \mathcal{M}(p)=1} \} \label{equ1} \\
    \mathcal{B}_i = \{p |_{ \exists p': \; p' \in \mathcal{O}_i \; \& \; ||p'-p||_{\infty}=1 \; \& \; \mathcal{M}(p)=0} \} \label{equ2}
\end{align}

However, a limitation of this straightforward boundary identification method is that the extracted boundaries may be non-convex, which can significantly prolong the trajectory optimization and reduce trajectory quality when non-convex obstacles are present~\cite{f1,f2}. To make boundaries as convex as possible, we segment the map $\mathcal{M}$ into multiple layers, with each layer $\mathcal{M}_i$ containing only a single group $\{\mathcal{O}_i, \mathcal{B}_i\}$. Using the group's maximum lengths in both x and y directions, a rectangular kernel filled with ones is applied to perform a closing operation on the group, resulting in an expanded map $\mathcal{M}_i'$ and a convex group $\{\mathcal{O}_i', \mathcal{B}_i'\}$ (see Fig.\ref{max_boundary}c).


However, $\mathcal{B}_i'$ is not directly usable due to the overlapping of obstacle groups (such as the overlap between groups with IDs $j$ and $k$ in the grey area of Fig.\ref{max_boundary}c). To address this, we define a concept of the corner set $\mathcal{C}_i$, which consists of points that consistently exist within both boundaries $\mathcal{B}_i$ and $\mathcal{B}_i'$, and have at least four collision-free points among their eight neighboring points (see Fig.\ref{max_boundary}c).
\begin{equation}
    \mathcal{C}_i = \{ p |_{p \in \mathcal{B}_i \; \& \; p \in \mathcal{B}_i' \; \& \; (\sum_{||p'-p||_{\infty}=1} \mathcal{M}_i'(p')=0)>=4  }  \}
\end{equation}

Traverse each corner set $\mathcal{C}_i$ sequentially. If the Bresenham connection~\cite{bresenham} between two adjacent corner points $c_i^j \in \mathcal{C}_i$ and $c_i^{j+1} \in \mathcal{C}_i$ does not intersect with any other boundary set or obstacle set, retain that portion of $\mathcal{B}_i'(c_i^j, c_i^{j+1})$. Otherwise, use the corresponding portion of $\mathcal{B}_i(c_i^j, c_i^{j+1})$ to obtain the max-convex boundary $\mathcal{B}_i^*$ (see Fig.\ref{max_boundary}d):
\begin{equation}
    \mathcal{B}_i^* = \{ \mathcal{B}_i'(c_i^j, c_i^{j+1}) \; or \; \mathcal{B}_i(c_i^j, c_i^{j+1}) |_{i \in [0, |\mathcal{C}_i|-1]}\}
\end{equation}

\begin{figure*}[thb]
    \centering
    \includegraphics[width=7.0in]{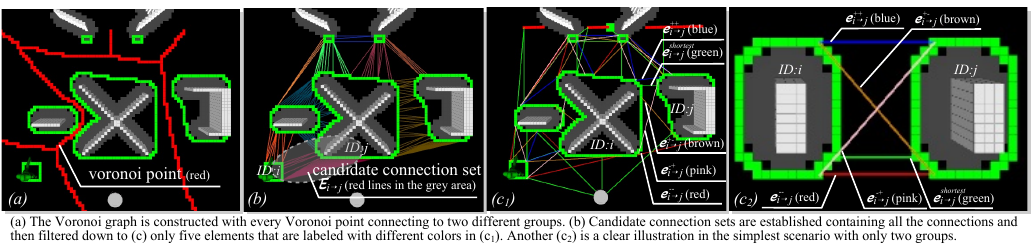}
    \caption{Illustration of group connections based on the group-level voronoi graph.}
    \label{connection}
\end{figure*}

\begin{figure}[thb]
    \centering
    \includegraphics[width=3.4in]{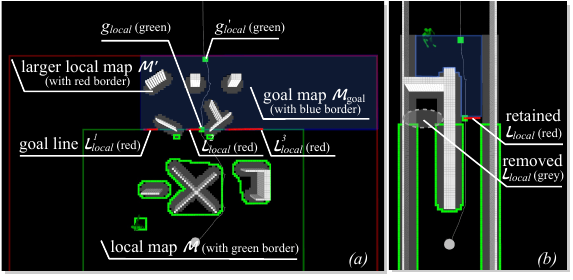}
    \caption{Goal lines in (a) lobby with all three goal lines being retained and (b) corridor with one goal line being removed.}
    \label{goal_lines}
\end{figure}

\subsection{Goal Lines Identification}
Assuming the intersection of the global path $\tau_{global}$ and the local map $\mathcal{M}$ serves as the local goal point $g_{local}$, a complete goal line can be initialized as $\mathcal{L}_{local}^-$, centered at $g_{local}$ and extending by a distance $D$ along the border of the local map $\mathcal{M}$. The length $D$ is proportional to the distance between the robot $p_{r}$ and the global goal $g_{global}$:
\begin{gather}
    \mathcal{L}_{local}^- = \{p|_{ ||p-g_{local}||_2<=D \; \& \; p \in border(\mathcal{M}_{local}) }\} \\
    D = min (\alpha || p_{r} - g_{global}||_2 , D_{max})
\end{gather}
where $\alpha$ is a scaling factor, and $D_{max}$ is a threshold. 

However, not all points on the goal line are usable, as some of them are occupied by obstacles, which divide the whole goal line into multiple small ones (see Fig.\ref{goal_lines}a):
\begin{gather}
    \mathcal{L}_{local} = \{\mathcal{L}_{local}^1, \mathcal{L}_{local}^2, ..., \mathcal{L}_{local}^n   \}
\end{gather}
Moreover, due to the presence of obstacles, some goal lines may lead to dead ends which should be removed (see Fig.\ref{goal_lines}b). To address this issue, we consider another local map $\mathcal{M}'$ that is larger than $\mathcal{M}$, with data derived from the overlay of the global map and local observations. Suppose the global path intersects $\mathcal{M}'$ at a point $g_{local}'$. The border of the quadrant where $g_{local}'$ is located, together with all goal lines $\mathcal{L}_{local}$, will form a goal map $\mathcal{M}_{goal}$. Subsequently, collision-free points are clustered like the process in Equ.\ref{equ1}, and the goal lines that belong to the same group of $g_{local}'$ will be retained:
\begin{gather}
    \mathcal{L}_{local}^* = \{ \mathcal{L}_{local}^i|_{ \mathcal{L}_{local}^i \in \mathcal{O}_j \; \& \; g_{local}' \in \mathcal{O}_k \; \& \;  j=k }   \}
\end{gather}

\subsection{Groups Connection Based on Group-level Voronoi Graph}
Taking the robot's current position $p_r$, group boundaries $\mathcal{B}_i^*$, and goal lines $\mathcal{L}_{local}^*$ as nodes, their connections $E_{i \to j}^*$ are defined in this subsection to form a topological map. 

For connections between groups, a Voronoi graph~\cite{voronoi} is constructed using all boundary points as input, with each Voronoi point $p$ being equidistant to the two nearest boundary points $p_1=V(p)$ and $p_2=V(p)$. The group attributes are utilized, meaning that only Voronoi points connecting different groups are retained (see Fig.\ref{connection}a). According to the Voronoi graph's properties, the two boundary points derived by any Voronoi point can always be connected without collision~\cite{voronoi2}. In this case, for two boundary $\mathcal{B}_i^*$ and $\mathcal{B}_j^*$, its candidate connection set $E_{i \to j}$ is initialized as (see Fig.\ref{connection}b): 
\begin{equation}
    E_{i \to j} = \{ \overrightarrow{p_1 p_2} |_{ p_1=V(p) \cap \mathcal{B}_i^* \; \& \; p_2 =V(p) \cap \mathcal{B}_j^* \; \& \; i \neq j}  \}
\end{equation}
The connection $E_{i \to j}$ can be further filtered down to five elements, which are subsequently used to construct trajectories. 

First, the shortest connection will be retained: 
\begin{equation}
    e_{i \to j}^{shortest} = \{ p_1 p_2 |_{argmin_{p_1p_2 \in E_{i \to j}} ||p_1-p_2||_2 }    \}
\end{equation}

Next, consider all points on the boundary $\mathcal{B}_i^*$ that can connect to $\mathcal{B}_j^*$. Define the first point encountered when traversing $\mathcal{B}_i^*$ clockwise as $p_{i \to j}^+$, and the first point encountered when traversing $\mathcal{B}_i^*$ counterclockwise as $p_{i \to j}^-$. For simplicity, the symbol \(+\) will denote clockwise, and \(-\) will denote counterclockwise in the following discussion. Similarly, identify \(p_{j \to i}^+\) and \(p_{j \to i}^-\) on the boundary $\mathcal{B}_j^*$. The properties of the Voronoi graph ensure that \(p_{i \to j}^+\) and \(p_{j \to i}^-\) are derived from the same Voronoi point~\cite{voronoi2}, making the connection \(e_{i \to j}^{++} = p_{i \to j}^+ p_{j \to i}^-\) collision-free. This connection physically represents the fastest switch from clockwise traversal around group $i$ to clockwise traversal around group $j$. Similarly, the fastest switch from counterclockwise to counterclockwise, $e_{i \to j}^{--}$, can also be obtained (see Fig.\ref{connection}c$_1$ and Fig.\ref{connection}c$_2$):
\begin{equation}
    e_{i \to j}^{++} = p_{i \to j}^+ p_{j \to i}^- \hspace{15pt} , \hspace{15pt}
    e_{i \to j}^{--} = p_{i \to j}^- p_{j \to i}^-
\end{equation}

\begin{figure*}[thb]
    \centering
    \includegraphics[width=7.0in]{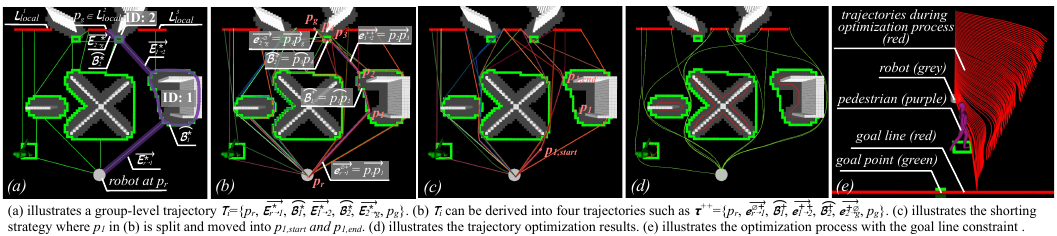}
    \caption{Illustration of trajectory initialization and optimization.}
    \label{trajectory_generation}
\end{figure*}

For the switch from clockwise to counterclockwise $e_{i \to j}^{+-}$, the ideal connection is $p_{i \to j}^+  p_{j \to i}^+$. However, as it might result in a collision, a shrinking strategy is required. Starting from $p_i = p_{i \to j}^+$ and $p_j = p_{j \to i}^+$, $p_i$ and $p_j$ is moved alternatively until the connection of them is collision-free. Similarly, starting from $p_i' = p_{i \to j}^-$ and $p_j' = p_{j \to i}^-$, the switch from counterclockwise to clockwise $e_{i \to j}^{-+}$ can be obtained: 
\begin{equation}
    e_{i \to j}^{+-} = p_{i} p_{j} \hspace{15pt} , \hspace{15pt}  
    e_{i \to j}^{-+} = p_{i}' p_{j}'
\end{equation}
Conclusively, the connection of two groups $E_{i \to j}^*$ consists of: 
\begin{equation}
    E_{i \to j}^* = \{ \overrightarrow{e_{i \to j}^{shortest}}, \;  \overrightarrow{e_{i \to j}^{++}}, \; \overrightarrow{e_{i \to j}^{--}}, \; \overrightarrow{e_{i \to j}^{+-}}, \; \overrightarrow{e_{i \to j}^{-+}} \}
\end{equation}

The connection between the robot $p_r$ and a boundary $\mathcal{B}_i^*$ is their shortest connection. Notably, some boundaries may not be connectable to the robot:
\begin{equation}
    E_{r \to i}^* = \{  \overrightarrow{p_r p_i} |_{p_i = argmin_{p_i \in \mathcal{B}_i^*} ||p_r-p_i||_2}  \}
\end{equation}


The connection between a boundary $\mathcal{B}_i^*$ and goal lines $\mathcal{L}_{local}^*$ is also their shortest connection, from $p_i \in \mathcal{B}_i^*$ to $p_g \in \mathcal{L}_{local}^j$. Notably, all goal lines are equivalent, meaning that a boundary will connect to only one goal line $\mathcal{L}_{local}^j$.
\begin{equation}
    E_{i \to g}^* = \{  \overrightarrow{p_i p_g} |_{p_i, p_g = argmin_{p_i \in \mathcal{B}_i^*, p_g \in \mathcal{L}_{local}^*} ||p_i-p_g||_2}  \}
\end{equation}






\section{Trajectory Initialization and Optimization}
\subsection{Group-level Trajectory Initialization}
A depth-first search is performed on the topological map to obtain the group-level trajectory set $\mathcal{T}$. Starting from the robot's position $p_r$, connectable groups are added to the open list, and the order of adding groups from $i$ to $j$ is determined by a heuristic value that considers both the connection length and the distance to the nearest local goal point $p_g$:
\begin{equation} 
  f_{i \to j} = || e_{i \to j}^{shortest} ||_2 + ||p_j - p_g||_2 
\end{equation}
where $p_j = \mathcal{B}_j^* \cap e_{i \to j}^{shortest}$, $p_g = argmin_{p_g \in \mathcal{L}_{local}^*} ||p_j p_g||_2$.

Two pruning strategies are designed to reduce unnecessary connections. The first is \textit{father\_visit}, which means that if a group $j$ can be connected by the father of $i$, the connection $i \to j$ will not be considered. The second is \textit{orientation\_limitation}, meaning that if the orientation of the shortest connection deviates significantly from the direction of the robot toward the local goal point, the connection $i \to j$ will not be considered. 
In this case, a group-level trajectory $\mathcal{T}_i$, consisting of a sequence of alternative connections $\overrightarrow{E_{i\to j}^*}$ and group detours $\wideparen{\mathcal{B}_i^*}$, can be described as (see Fig.\ref{trajectory_generation}a):
\begin{equation}
    \mathcal{T}_i = \{ p_r, \overrightarrow{E_{r \to 1}^*}, \wideparen{\mathcal{B}_1^*}, \overrightarrow{E_{1 \to 2}^*}, \wideparen{\mathcal{B}_2^*}, ..., \overrightarrow{E_{2 \to g}^*}, p_g  \}
\end{equation}

\subsection{Trajectories Derivation from Group-level Trajetories}
The detour around a group $\wideparen{\mathcal{B}_i^*}$ can be detailed as either clockwise $\wideparen{\mathcal{B}_i^+}$ or counterclockwise $\wideparen{\mathcal{B}_i^-}$. Thus, a group-level trajectory consisting of $n$ groups can quickly derive $2^n$ trajectories. Once the detour directions for adjacent groups $i$ and $i+1$ are determined, their connection $\overrightarrow{E_{i \to i+1}^*} = \overrightarrow{e_{i \to i+1}^{\circ \circ}}$ can be immediately established, representing the switch from the detour around group $i$ to that around group $i+1$, where the first ${\circ \in \{+,-\}}$ corresponds to the direction of $\wideparen{\mathcal{B}_i^*}$, and the second $\circ$ corresponds to the direction of $\wideparen{\mathcal{B}_{i+1}^*}$.
The connection from the robot \( p_r \) to group \( i \), $\overrightarrow{e_{r\to i}^{\varnothing \circ}}$, is more straightforward. Assuming their shortest connection $\overrightarrow{e_{r \to i}^{shortest}}$ intersects the boundary $\mathcal{B}_i^*$ at \( p_i \), \( p_i \) moves along the boundary in the detour direction \( \overrightarrow{\mathcal{B}_i^*} \) until the Bresenham connection between \( p_r \) and \( p_i \) is going to intersect an obstacle point.
Additionally, the connection from a group \( i \) to the goal lines $\mathcal{L}_{local}^*$, $\overrightarrow{e_{i \to g}^{\circ \varnothing}}$, starts at the intersection point \( p_i \) of the connection $\overrightarrow{e_{i-1 \to i}^{\circ \circ}}$ with the boundary \( \mathcal{B}_i^* \). From there, \( p_i \) moves in the opposite direction of \( \overrightarrow{\mathcal{B}_i^*} \), searching for the nearest goal point \( p_g \). This process continues until \( \overrightarrow{p_ip_g} \) is going to encounter an obstacle or boundary point. Notably, the connected goal point and goal line may differ from the group-level goal (see Fig.\ref{trajectory_generation}b).

However, although the boundary $\wideparen{\mathcal{B}_i^*}$ is max-convex and the connection $\overrightarrow{e_{i \to i+1}^{\circ \circ}}$ represents the nearly fastest switch, unnecessary detours can still occur at the intersection \(p_i\) between the boundary and the connection. Therefore, a shorting strategy is required.
Starting from \(p_{i}^{\mathcal{B}}\)=\(p_i\) and \(p_{i}^{E}\)=\(p_i\), the two points incrementally move along the boundary and connection respectively until their connection encounters an obstacle point. This additional line segment \(p_{i}^{\mathcal{B}} p_{i}^{E}\) helps reduce unnecessary detours in the trajectory (see the comparison of the pink trajectory between Fig.\ref{trajectory_generation}b and Fig.\ref{trajectory_generation}c).

All trajectories are sorted in ascending order by length, and their H-signature values~\cite{e1} are checked. Trajectories with the same H-signature but longer lengths are discarded, while the others are retained for further optimization (see Fig.\ref{trajectory_generation}d).

\subsection{Trajectory Optimization with Goal Line Constraints}
The trajectory optimization process follows our previously proposed incremental optimization framework~\cite{e5}. Initially, multiple trajectories are initialized, assuming that the obstacles and pedestrians are static. Then, their speeds are gradually increased and the trajectory is optimized. 

The introduction of the goal line makes the end of the trajectory $p_n$ not fixed to a specific point during the optimization process, but rather to an entire goal line $\mathcal{L}_{local}^k$ that $p_n$ belongs to (See Fig.\ref{trajectory_generation}e). 
For a goal line with two ends $p_1$ and $p_2$, $p_n$ should satisfy the constraint: 
\begin{equation}
    g(p_n) = ||p-p_1||_2 + ||p-p_2||_2 - ||p_1-p_2||_2  =0
\end{equation}

\begin{figure*}[thb]
    \centering
    \includegraphics[width=7.0in]{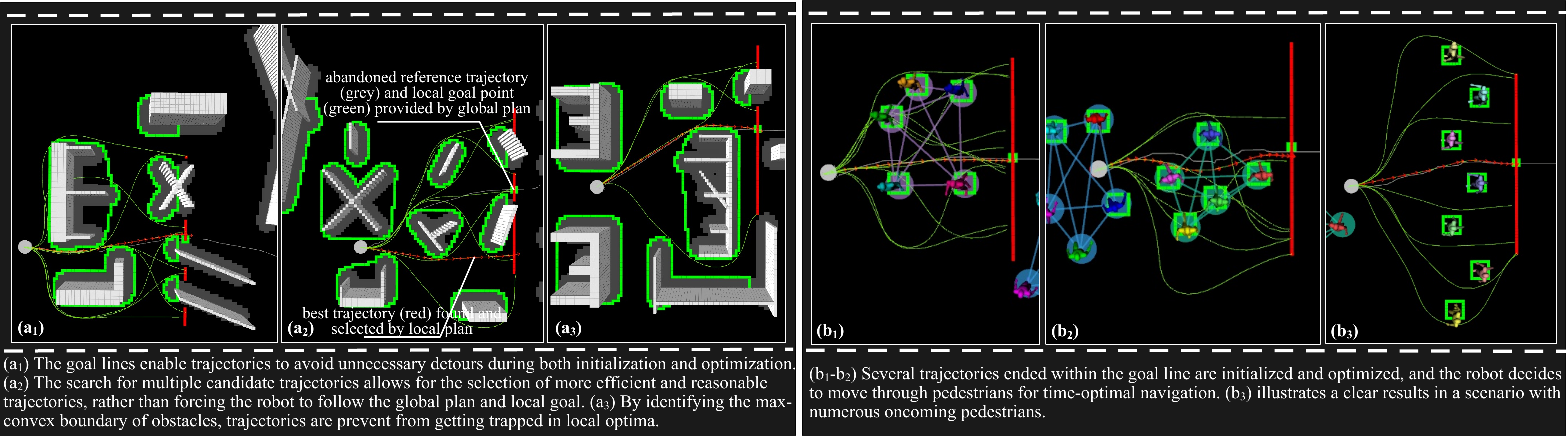}
    \caption{Performance of GA-TEB in static and dynamic scenes. Visit the \href{https://ga-teb.github.io}{website} for video demonstrations and source code. }
    \label{sim1}
\end{figure*}

\section{Simulation}
The simulation evaluates the performance of the proposed GA-TEB framework in both complex static scenarios with multiple non-convex obstacles and dynamic scenarios involving numerous pedestrians. Performance is measured across four metrics: trajectory initialization time, trajectory optimization time, time to reach the global goal, and success rate. The results are compared against three relevant methods: Voronoi-TEB~\cite{e3}, Ego-TEB~\cite{e4}, and Graphic-TEB~\cite{e5}.

\begin{table}[]
\centering
\caption{Comparison Results}
\begin{tabular}{lcccccc}
\hline
\multicolumn{1}{c}{\multirow{2}{*}{}} & \multirow{2}{*}{$T_{init}$ {[}s{]}} & \multirow{2}{*}{$T_{opt}$ {[}s{]}} & \multicolumn{2}{c}{$T_{goal}$ {[}s{]}} & \multicolumn{2}{c}{$R_{goal}$ {[}\%{]}} \\ \cline{4-7} 
\multicolumn{1}{c}{}                  &                                &                               & static          & dyna.         & static          & dyna.          \\ \hline
Voronoi                    & 0.304                 & 0.182                         & 36.2            & 40.4            & 94              & 60               \\ \hline
Ego                           & -                     & -                             & 33.5            & 37.4            & 100             & 78               \\ \hline
Graphic                         & 0.029                 & 0.143                         & 34.1            & 38.0            & 98              & 84               \\ \hline
GA (ours)                         & \textbf{0.009}                 & \textbf{0.054}                         & \textbf{32.0}            & \textbf{33.2}            & \textbf{100}             & \textbf{98}               \\ \hline
\end{tabular}
\label{table}
\end{table}

In the simulation, the robot model used is a differential-drive robot with a maximum speed of 1 m/s and a radius of 0.3 m. Static obstacles vary in shape—lines, rectangles, circles, T-shapes, and X-shapes—with randomly assigned positions and sizes. Pedestrians, driven by the social force model~\cite{okal2014towards} (ignoring the force from the robot to make the test more strict), have a maximum speed of 0.5 m/s and a radius of 0.3 m, with randomly generated initial positions and velocities. A total of 20 random scenarios are generated for both static and dynamic environments, with data recorded during the robot’s movement. The average performance results are presented in Table.\ref{table}. Further details regarding the simulation setup are available in the source code\footnote{https://ga-teb.github.io}.

\textbf{Trajectory initialization time ($T_{init}$)}. 
The time required for trajectory initialization primarily stems from the depth-first search and establishing connections between nodes. We benchmark the number of trajectories obtained by Voronoi-TEB as the complete set~\cite{e1}, which is widely accepted. Due to its short-sightedness, Ego-TEB finds only about half of the trajectories and is therefore excluded from this discussion. Both Graphic-TEB and our GA-TEB (without applying \textit{orientation\_limitation}) can find complete trajectories.

\begin{figure}[thb]
    \centering
    \includegraphics[width=3.4in]{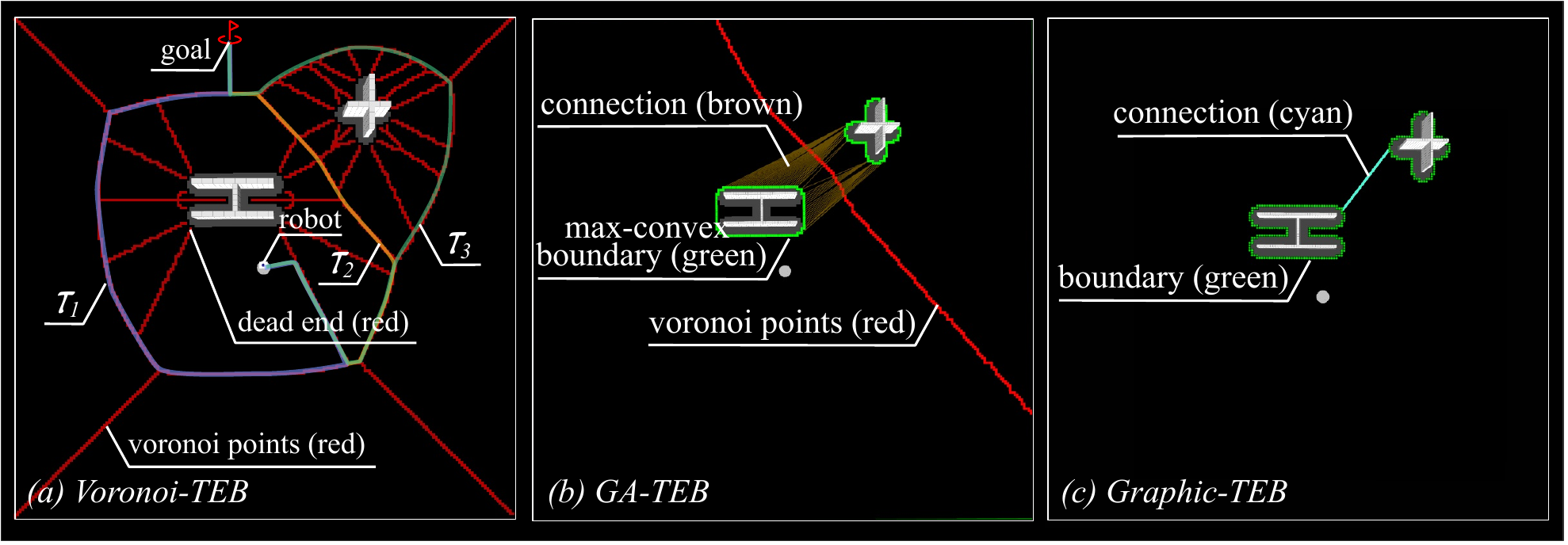}
    \caption{Comparison of initializing and optimizing trajectories. }
    \label{sim2}
\end{figure}

The comparison with Voronoi-TEB focuses on the time spent in depth-first search. Voronoi-TEB does not account for group attributes, resulting in three issues (see the comparison in Fig.\ref{sim2}a-b): first, it extracts many irrelevant Voronoi points, leading to inefficient searches that often end in dead-ends; second, the search operates over adjacent points rather than nodes, as in GA-TEB, which reduces efficiency; third, it cannot leverage pruning strategies such as \textit{orientation\_limitation} to remove unnecessary search in practical application. Additionally, the concept of group-level trajectories introduced in GA-TEB allows for the rapid derivation of $2^n$ ordinary trajectories, further accelerating the trajectory generation.

The comparison with Graphic-TEB centers on connection establishment. Graphic-TEB compares all boundary points of adjacent groups to find the shortest connection between them, with a time complexity of $O(n^2)$. This becomes extremely time-consuming in environments with densely distributed, non-convex obstacles. In contrast, GA-TEB introduces a Voronoi graph aligned with group-level attributes, reducing the time required by about two-thirds.

\textbf{Trajectory optimization time ($T_{opt}$)}. 
The time required for trajectory optimization is primarily influenced by the quality of the initial trajectory. We define the optimization time as the sum of all point position changes between two consecutive optimizations being less than 0.1m. In Voronoi-TEB, the trajectory consists of Voronoi points which are the midpoints between obstacles (see Fig.\ref{sim2}a with three initialized trajectories \{$\tau_1,\tau_2,\tau_3$\} ). It is a conservative strategy that requires significant time to optimize and shorten the trajectory length. In contrast, Graphic-TEB initializes its trajectory points very close to obstacles, which leads to prolonged optimization times in environments with non-convex obstacles (see Fig.\ref{sim2}c with the green boundary), as the trajectory needs time to detach from the obstacles.

GA-TEB keeps a balance between them, mitigating the influence of non-convex obstacles by leveraging the max-convex boundaries (see Fig.\ref{sim2}b with green boundary) and the fast switch connections (see Fig.\ref{sim2}b for all the candidate connections between the adjacent obstacles), resulting in a great reduction in optimization time.

\textbf{Time to the global goal ($T_{goal}$) and success rate ($R_{goal}$)}. 
We evaluate these two metrics in both static and dynamic scenarios where the distance between the start and goal points exceeds 30 meters, and our method performs the best. 

Fig.\ref{sim1}(a) illustrates the reasons behind GA-TEB's strong performance in static scenarios with non-convex obstacles. (a$_1$) The introduction of multiple goal lines allows trajectories to connect directly to the nearest goal point, avoiding unnecessary turns during initialization and optimization. (a$_2$) The provision of multiple candidate trajectories, each ending within the goal line, enables the generation and selection of more efficient and reasonable trajectories. While the reference trajectory provided by the global plan requires the robot to make a sharp turn, forcing it to slow down, GA-TEB offers alternative options that allow the robot to move forward smoothly and efficiently. (a$_3$) In scenarios with non-convex obstacles, GA-TEB ensures their boundaries are as convex as possible, accelerating trajectory optimization and preventing the optimization from getting trapped in local optima.

Fig.\ref{sim1}(b) exemplifies the dynamic scenarios with pedestrians. (b$_1$) and (b$_2$) illustrate two episodes where the crowds driven by the social force model move together and the robot decides to move through crowds for higher efficiency. (b$_3$) illustrates the clear results obtained by GA-TEB in the situation with oncoming pedestrians. The highest success rate of GA-TEB can be attributed to the fast trajectory initialization and optimization, which enhances the robot’s real-time reaction, particularly in this kind of dynamic environment with a large number of pedestrians. Additionally, the introduction of the goal line prevents the freezing problem and thus reduces $T_{goal}$. In 50 navigation tests, the total number of freezing incidents for Voronoi-TEB, Graphic-TEB, Ego-TEB, and GA-TEB are 112, 49, 33, and 0, respectively.





%


\begin{figure*}[thb]
    \centering
    \includegraphics[width=7.0in]{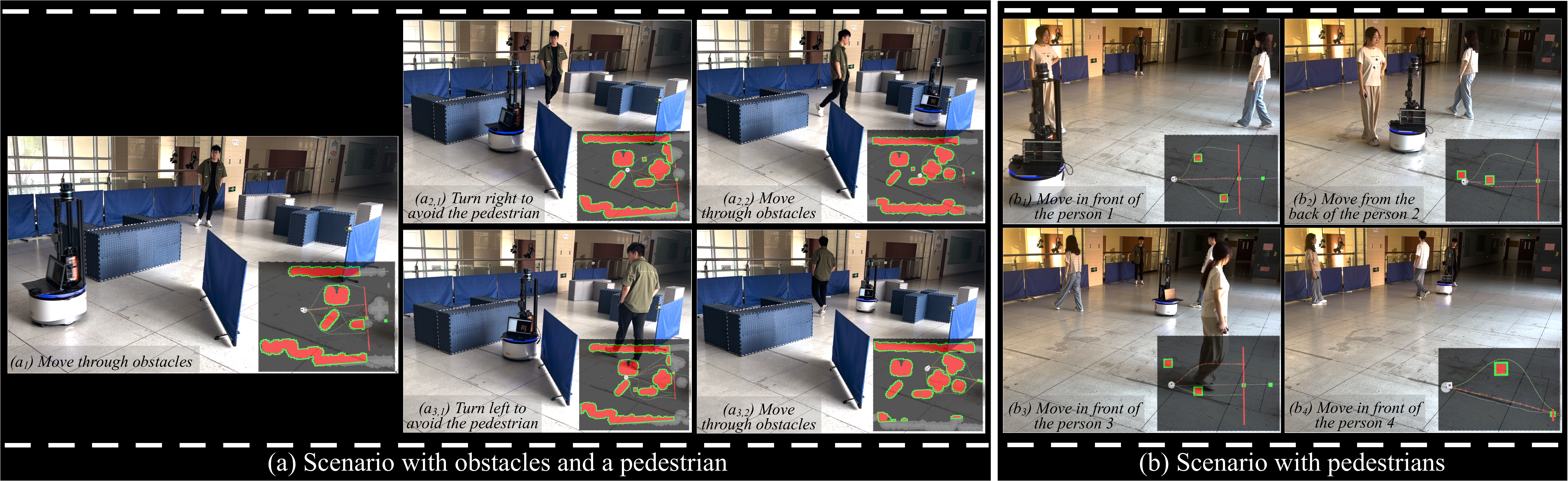}
    \caption{Performance of GA-TEB in the experiment. (a$_1$) The robot navigates through obstacles and turns (a$_{2,1}$-a$_{2,2}$) right or (a$_{3,1}$-a$_{3,2}$) left to avoid the pedestrian. (b$_1$-b$_4$) The robot moves in front of or behind pedestrians based on their relationship.}
    \label{exp}
\end{figure*}
\section{Experiment}
The experiments are conducted in scenarios involving both obstacles and pedestrians, as well as in scenarios with multiple pedestrians. The robot model used is similar to the one in the simulation, with an additional 16-line LiDAR mounted on top for pedestrian detection and tracking. 

Fig.\ref{exp}(a) highlights key episodes of the scenario with obstacles and pedestrians, with real-world scenes occupying the main area and the rviz board located in the bottom right. (a$_1$) The robot extracts the max-convex boundary of the nearest E-shaped obstacle and selects the shortest trajectory that navigates through the obstacles. (a$_{2,1}$) When encountering an oncoming pedestrian, the robot prefers to turn right to avoid a collision and then (a$_{2,2}$) smoothly continues through the obstacles ahead. In another situation, where (a$_{3,1}$) a pedestrian approaches from the robot’s rear right, the robot turns left to move in front of the pedestrian and then (a$_{3,2}$) navigates through the obstacles to proceed toward the goal.

Fig.\ref{exp}(b) illustrates the scenario with multiple pedestrians. When encountering two pedestrians moving perpendicular to the robot's direction, the robot chooses to navigate either (b$_1$) in front of or (b$_2$) behind them to ensure safety. (b$_3$) A pedestrian initially walks at a slow pace, prompting the robot to turn left to pass in front of him. As the pedestrian's speed increases, the robot quickly adjusts by turning at a sharper angle to avoid a collision. (b$_4$) When a fast-moving pedestrian intentionally blocks the robot's intended path, the robot robustly switches its strategy to ensure safety.

\section{Conclusion}
Focusing on crowd navigation, this paper introduces the concept of \textit{goal lines}, which extends the traditional goal point to multiple lines, preventing the robot from experiencing freezing issues and minimizing unnecessary detours during trajectory optimization. In conjunction with goal lines, a topological map construction strategy is proposed to model obstacle groups as convex as possible and establish their connections using five basic elements, enabling both rapid and high-quality trajectory initialization.
Simulations and experiments are conducted in static scenarios with non-convex obstacles and dynamic scenarios involving numerous pedestrians. The proposed method, GA-TEB, demonstrates the ability to quickly initialize and optimize trajectories, allowing the robot to respond promptly in complex environments and achieve the highest success rate. Furthermore, by avoiding unnecessary detours during optimization, the method prioritizes more efficient trajectories, enabling the robot to reach the global goal more effectively.

\bibliographystyle{IEEEtran}
\bibliography{./bibtex/IEEEabrv}
\end{document}